\def\year{2019}\relax
\documentclass[letterpaper]{article} % DO NOT CHANGE THIS
\usepackage{aaai19}  % DO NOT CHANGE THIS
\usepackage{times}  % DO NOT CHANGE THIS
\usepackage{helvet} % DO NOT CHANGE THIS
\usepackage{courier}  % DO NOT CHANGE THIS
\usepackage[hyphens]{url}  % DO NOT CHANGE THIS
\usepackage{graphicx} % DO NOT CHANGE THIS
\usepackage{bm,algorithm,algorithmic}

\newtheorem{examp}{Example}
\newtheorem{defi}{Definition}

\usepackage{amsmath}
\DeclareMathOperator*{\argmax}{arg\,max}
\DeclareMathOperator*{\argmin}{arg\,min}
\usepackage{graphicx,tikz,pgfplots}  % DO NOT CHANGE THIS
\title{Approximate MMAP by Marginal Search}
\author{Alessandro Antonucci\\IDSIA\\Lugano (Switzerland)\\{\tt alessandro@idsia.ch}
\And
Thomas Tiotto\\Groningen Cognitive Systems and Materials\\
Groningen (The Netherlands)\\{\tt t.f.tiotto@rug.nl}}
\begin{document}
\maketitle
\begin{abstract}
We present a heuristic strategy for marginal MAP (MMAP) queries in graphical models. The algorithm is based on a reduction of the task to a polynomial number of  marginal inference computations. Given an input evidence, the marginals mass functions of the variables to be explained are computed. Marginal information gain is used to decide the variables to be explained first, and their most probable marginal states are consequently moved to the evidence. The sequential iteration of this procedure leads to a MMAP explanation and the minimum information gain obtained during the process can be regarded as a confidence measure for the explanation. Preliminary experiments show that the proposed confidence measure is properly detecting instances for which the algorithm is accurate and, for sufficiently high confidence levels, the algorithm gives the exact solution or an approximation whose Hamming distance from the exact one is small.
\end{abstract}

\section{Introduction}
Probabilistic graphical models such as Bayesian networks and Markov random fields are popular tools for a compact generative description of the uncertain relations between the variables in a system \cite{koller2009}. Reasoning with such models is achieved by inferential computations involving sums and maximizations among the local components (potentials or conditional probability tables).

Typical inference tasks in these models can be regarded as special cases of a general task called \emph{marginal MAP} (MMAP). In a MMAP task a set of model variables should be \emph{explained}, i.e., their joint most probable state should be detected, while some of the other variables are \emph{observed} in a given state, and the remaining ones should be \emph{marginalized}, i.e., summed out. Complexity analysis reveals that MMAP is a NP$^{\mathrm{PP}}$-complete \cite{park2004}. Notable MMAP sub-cases correspond to situations in which: (i) there are no variables to explain and the problem corresponds to the computation of the probability of the observed variables; and (ii) there are no variables to marginalize and the problem is to find the most probable state of the variables to explain given an observation of all the other variables. The complexity of these two tasks, sometimes called, respectively, PR and MAP inference, is lower as PR is \#P-complete and MAP is NP-complete. In practice MMAP is a much harder task than PR or MAP and, for instance, for singly-connected topologies polynomial solutions of PR and MAP can be derived while MMAP remains NP-hard \cite{koller2009}. Despite such high complexity, as noticed in \cite{marinescu2018stochastic}, MMAP is a very important task, as it corresponds to the case of a model with latent variables, which are commonly used in graphical models to express non-trivial dependency patterns. Various \emph{anytime} algorithms providing lower and upper bounds to the optimal MMAP values have been proposed, e.g., \cite{maua2012}, and the state of the art is currently the bounding scheme based on stochastic search proposed on \cite{marinescu2018stochastic}.

\emph{Marginal inference} (MAR) is a third important MMAP sub-case for which only a single variable is explained. It is straightforward to reduce MAR to a number of calls to PR equal to the number of states of the variable to explain and its complexity remains therefore \#P-complete. In this paper we reduce MMAP to a polynomial number of MAR calls. Given the evidence of the MMAP task, our procedure uses MAR to compute the marginal mass functions of the variables to explain and ``move'' to the observation the variables with the most extreme probabilistic values. The iteration of this procedure represents a heuristic approach to approximate MMAP. Different  information-theoretic criteria can be considered to drive such a search for the most probable configuration in order to define scores to characterize the reliability of the corresponding explanation. These scores can be also used to decide when the procedure should be terminated, thus providing a partial-but-reliable MMAP explanation. The paper is organized as follows: we first review the existing work in the field and formalize the problem with the necessary notation, the heuristic strategy together and the scores are consequently described, and an empirical validation reported together with a discussion about relations with existing work and possible outlooks is finally provided.

\section{Related Work}
In \cite{Butz2018}, a procedure similar to the one presented in this paper has been considered in the context of explainable AI and Bayesian networks. Rather than focusing on the algorithmic task, the goal of that procedure is to generate a linguistic explanation of the input evidence and a description of the reasoning process behind the model. Here we make explicit how such scheme would not necessarily return exact explanations and provide an information-theoretic score able to characterize their  confidence level, this being proved to be more effective than the highest probability level considered in that paper.

Concerning the MMAP literature, most of the existing algorithms are exact schemes possibly giving anytime approximations \cite{park2002,maua2012,marinescu2014,marinescu2018stochastic}. Variational methods reducing the task to message propagation have been proposed instead for approximate inference \cite{jiang2011,liu2013}. %Alternatively, recent approaches approximate the intractable counting task involved by the marginalization in MMAP to a set of constrained optimization tasks \cite{xue2016}. This has some analogy with what is done in the present paper, where the counting is implicitly transferred to the MAR tasks.

\section{Background}
We consider discrete random variables only. If $X$ is a variable, the finite set $\Omega_X$ denotes its set of possible values, $|\Omega_X|$ is the cardinality of this set, and $x$ is a generic element of $\Omega_X$. A probability mass function $P$ is a non-negative real-valued map defined over $\Omega_X$ and normalized to one. For each $x\in\Omega_X$, $P(x)$ is the probability for $X=x$. The entropy of a mass function $P$ over $X$ is defined as 
$H[P(X)]:=-\sum_{x\in\Omega_{X}} P(x) \log_{|\Omega_{X}|} P(x)$. Note that $H[P(X)]\in[0,1]$, being one with uniform mass functions and zero in the degenerate case of probabilities equal to zero and one. Given a joint variable $\bm{X}:=(X_1,\ldots,X_n)$, we can similarly define a joint mass function $P(\bm{X})$.
Given $\bm{x}\in\Omega_{\bm{X}}$ and $\bm{X}'\subset\bm{X}$, notation $\bm{x}^{\downarrow \bm{X}'}$ is used for the restriction to the variables in $\bm{X}'$ of the states in $\bm{x}$. A potential $\phi$ is just an un-normalized (but still non-negative) mass function. Say that for each $i=1,\ldots,f$, $\phi_i$ is a potential over $\bm{X}_i \subset \bm{X}$ and such that $\cup_{i=1}^f \bm{X}_i = \bm{X}$. If this is the case we call $\bm{\Phi}:=\{\phi_i\}_{i=1}^f$ a graphical model (GM) over $\bm{X}$. A GM defines a joint mass function over $\bm{X}$ such that $P(\bm{X}) \propto \prod_{i=1}^f \phi_i(\bm{X}_i)$. Note that both Bayesian networks and Markov random fields can be regarded as GMs. We are now in the condition of defining MMAP inference in GMs.

\begin{defi}[MMAP]
Given a GM $\bm{\Phi}$ over $\bm{X}$, the partition $(\bm{X}_M,\bm{X}_S,\bm{X}_E)$ of $\bm{X}$, and an \emph{observation} $\bm{X}_E=\bm{x}_E$, a MMAP task consists in the computation of state:
\begin{equation}
\bm{x}_M^* := \argmax_{\bm{x}_M\in\Omega_{\bm{X}_M}} \sum_{\bm{x}_S\in\Omega_{\bm{X}_S}} P(\bm{x}_M,\bm{x}_S,\bm{x}_E)\,,
\end{equation}
and the corresponding probability $p^*:=P(\bm{x}_M^*,\bm{x}_E)$.
\end{defi}
If $\bm{X}_M=\emptyset$, MMAP is called PR and it only consists in the computation of $P(\bm{x}_E)$. Although both problems are NP-hard in general, PR is considerably simpler task (being \#P-complete) compared to general (NP$^\mathrm{PP}$-complete) MMAP. Let us also define the MAR task.
\begin{defi}[MAR]
Given a GM $\bm{\Phi}$ over $\bm{X}$, an observation $\bm{x}_E$ of the variables $\bm{X}_E\subset \bm{X}$, and a single variable $X \in \bm{X}\setminus \bm{X}_E$, a MAR task consists in the computation of $P(x|\bm{x}_E)$ for each $x\in\Omega_{X}$.
\end{defi}
It is straightforward to solve MAR by using PR to compute $P(x,\bm{x}_E)$ for each $x\in\Omega_{X}$, as the normalization of these joint probabilities gives the MAR conditional probabilities.

\section{Approximating MMAP by Multiple MARs}
As noticed in the previous section MMAP becomes simpler if $\bm{X}_M$ contains a single variable. Yet, as shown by the following example from \cite{liu2013}, MMAP cannot be trivially reduced to a sequence of MAR tasks.

\begin{examp}[Wheather Dilemma]\label{ex:w}
Variable $R$ and $D$ denote, respectively, whether or not it is a rainy day in Irvine, and whether or not Alice is going to the office by car. Accordingly let us assume $\Omega_D:=\{rainy,sunny\}$ and $\Omega_R:=\{walk,drive\}$. The assessments for the marginal probability $P(rainy)=0.4$ and the conditional probabilities $P(drive|rainy)=.875$ and $P(drive|sunny)=0.5$ are sufficient to compute the joint mass function $P(R,D)$ displayed in Table 
\ref{tab:irvine}. State $sunny$ is the one with the highest marginal probability for $R$ and, similarly, $drive$ has the highest marginal for $D$. Yet, the most probable joint state of $(R,D)$ is $(rainy,drive)$ (see bold numbers in Table \ref{tab:irvine}).
\end{examp}
\begin{table}[htp!]
\centering
\begin{tabular}{ccccc}
\hline
$r$&$d$&$P(r,d)$&$P(r)$&$P(d)$\\
\hline
$sunny$&$walk$&$0.30$&$\mathbf{0.60}$&$0.35$\\
$rainy$&$walk$&$0.05$&$0.40$&-\\
$sunny$&$drive$&$0.30$&-&$\mathbf{0.65}$\\
$rainy$&$drive$&$\mathbf{0.35}$&-&-\\
\hline
\end{tabular}
\caption{Joint and marginal probabilities for Example \ref{ex:w}}
\label{tab:irvine}
\end{table}

The above example shows that a most probable joint configuration is not necessarily a combination of most probable marginal configurations. This is perfectly acceptable for non-independent variables. If we regard the identification of the most probable joint state as a MMAP task and the identification of the two most probable marginal states as a MAR task, Example \ref{ex:w} shows that MMAP cannot be trivially reduced to a sequence of MAR tasks over the variables to explain. Yet, in the following example we show that a more sophisticated scheme could be more effective in achieving such reduction.

\begin{examp}[Solving the Wheather Dilemma]\label{ex:w2}
Consider the same setup as in Example \ref{ex:w}. As shown in Table \ref{tab:irvine}, the marginal mass functions of the two variables are $P(R)=[.6,.4]$ and $P(D)=[.35,.65]$. Among these two mass functions, $P(D)$ is the one with the most extreme value, i.e., $drive$. We regard such an ``extreme'' state as an evidence and, consequently, compute $P(R|drive)=[6,7]/13$. Thus, the most probable (conditional) state of $R$ is $rainy$. In other words, the most probable joint state of the two variables is  the combination of the most probable marginal state for the variable with the most extreme marginal probability combined the most probable posterior state for the other variable after promoting the first to an evidence.
\end{examp}

Example \ref{ex:w2} suggests a heuristic strategy for MMAP tasks. In that example, the most probable configuration of the two variables is obtained sequentially by first explaining a variable, whose most probable state is promoted to evidence, and finally explaining the other. Note that starting from the variable with the most extreme values might be important, e.g., explaining $R$ before $D$ leads to a wrong conclusion.

The most extreme probabilistic value in the example can be intended as a proxy for the most informative (i.e., least entropic) mass function. The difference between the two descriptors might be important only when comparing mass functions over variables with different cardinality. Consider for instance the ternary mass function $P(X'):=[.2,.1,.7]$ and the binary mass function $P(X''):=[.8,.2]$, while $\max_{X'\in\Omega_{X'}} P(X')> \max_{X''\in\Omega_{X''}} P(X'')$ we have $H[P(X')]<H[P(X'')]$. When comparing binary variables the two descriptors are instead equivalent, as the entropy is a monotone function of the probability of the most probable state. With more than two states, again having the highest most probable state might not imply that the entropy is lower, e.g., $H[[0.75,0.24,0.01]]<H[[0.80,0.10,0.10]]$.

We are now in the condition of presenting our heuristic reduction of MMAP to MAR. This is detailed in Algorithm \ref{alg:mar2mmap}. Given a MMAP instance in input (line 1), a copy of the variables to be explained, the observed ones and their states are stored (line 2). The procedure consists in computing the MAR for each variable to be explained (lines 4-6), find the least entropic one (line 7), and its most probable state (line 8). Such a variable-state pair is moved to the evidence (lines 9-11). The procedure is iterated until all the variables to be explained are moved to the evidence (line 3). The resulting explanation is the restriction to the variables to be explained of the  evidence generated in this way (line 13).

\begin{algorithm}[htp]
\caption{MMAP2MAR}
\label{alg:mar2mmap}
\begin{algorithmic}[1]
\STATE {\bf input:} $(\bm{X}_M,\bm{X}_E,\bm{x}_E)$
\STATE $(\bm{X_M}',\bm{X_E}',\bm{x_E}') \leftarrow (\bm{X}_M,\bm{X}_E,\bm{x}_E)$
\WHILE{$\bm{X}_M' \neq \emptyset$}
\FOR{$X \in \bm{X}_M'$}
\STATE compute $P(X|\bm{x}_E')$
\ENDFOR
\STATE{${X}^* \leftarrow \argmin_{X\in\bm{X}_M'} H[P(X|\bm{x}_E')]$}
\STATE{$\tilde{x}^* \leftarrow \argmax_{x^* \in \Omega_{X^*}} P(x^*|\bm{x}_E')$}
\STATE{$\bm{X}_M' \leftarrow \bm{X}_M' \setminus \{ X^*\}$}
\STATE{$\bm{X}_E' \leftarrow \bm{X}_E' \cup \{ X^*\}$}
\STATE{$\bm{x}_E' \leftarrow \bm{x}_E' \cup \{ \tilde{x}^*\}$}
\ENDWHILE
\STATE {\bf output:} $\bm{x}_M^* \leftarrow$ $\bm{x}_E'^{\downarrow\bm{X}_M}$
\end{algorithmic}
\end{algorithm}

Overall, Algorithm \ref{alg:mar2mmap} requires a polynomial number of MAR calls, this number being clearly quadratic with respect to the cardinality of $\bm{X}_M$. In order to understand the kind of approximation induced by Algorithm \ref{alg:mar2mmap}, let us consider just for the sake of readability a simpler (MAP) task with both $\bm{X}_E$ and $\bm{X}_S$ empty and three variables to be explained, i.e., $p^*:=\max_{x_1,x_2,x_3}P(x_1,x_2,x_3)$. By considering the \emph{chain rule} with the natural order over the three variables, we have:
\begin{equation}\label{eq:outer}
p^*=\max_{x_1,x_2,x_3} P(x_3|x_2,x_1) P(x_2|x_1) P(x_1)\,,
\end{equation}
while, assuming that the order induced by the least entropic marginals is the natural one, Algorithm \ref{alg:mar2mmap} returns:
\begin{equation}\label{eq:tilde}
\tilde{p}:= \max_{x_3} P(x_3|\tilde{x}_2,\tilde{x}_1)
\max_{x_2} P(x_2|\tilde{x}_1)
\max_{x_1} P(x_1)\,,
\end{equation}
where $\tilde{x_1}$ is the $\argmax_{x_1} P(x_1)$ and 
$\tilde{x_2}$ is $\argmax_{x_2} P(x_2|\tilde{x}_1)$. If we also set $\tilde{x_3}:=\argmax_{x_3} P(x_3|\tilde{x_1},\tilde{x_2})$, by chain rule and Equation \ref{eq:tilde} we have $\tilde{p}=P(\tilde{x}_1,\tilde{x}_2,\tilde{x}_3)$ and thus, by Equation \eqref{eq:outer}, $p^* \geq \tilde{p}$. The result, which remains valid for general MMAP instances, says that in general Algorithm \ref{alg:mar2mmap} gives a lower bound for MMAP tasks.

After any iteration of the while loop in Algorithm \ref{alg:mar2mmap}, $H[P(X^*|\bm{x}_E)]$ can be intended as a confidence measure of the heuristic action of moving $(X^*=\tilde{x}^*)$ to the evidence. Algorithm \ref{alg:mar2mmap2} depicts a more cautious version of Algorithm \ref{alg:mar2mmap} that moves a most probable state to the evidence only if the minimum entropy of the marginal mass functions is below a threshold $\epsilon$ (line 8). If this is not the case, the iteration ends (line 14). After termination the values of the explained variables can be extracted from the evidence. Yet, unlike Algorithm \ref{alg:mar2mmap}, in this case only the variables of $\bm{X}_M$ not in $\bm{X}_M'$ are explained (line 17).

\begin{algorithm}[htp]
    \caption{$\epsilon$-MMAP2MAR}
    \label{alg:mar2mmap2}
    \begin{algorithmic}[1]
        \STATE {\bf input:} $(\bm{X}_M,\bm{X}_E,\bm{x}_E,\epsilon)$
        \STATE $(\bm{X_M}',\bm{X_E}',\bm{x_E}') \leftarrow (\bm{X}_M,\bm{X}_E,\bm{x}_E)$
        \WHILE{$\bm{X}_M' \neq \emptyset$}
        \FOR{$X \in \bm{X}_M'$}
        \STATE compute $P(X|\bm{x}_E')$
        \ENDFOR
        \STATE{$X^* \leftarrow \argmin_{X\in\bm{X}_M'} H[P(X|\bm{x}_E')]$}
        \IF{$H[P(X^*|\bm{x}_E')]<\epsilon$}
        \STATE{$\tilde{x}^* \leftarrow \argmax_{x^* \in \Omega_{X^*}} P(x^*|\bm{x}_E')$}
        \STATE{$\bm{X}_M' \leftarrow \bm{X}_M' \setminus \{ X^*\}$}
        \STATE{$\bm{X}_E' \leftarrow \bm{X}_E' \cup \{ X^*\}$}
        \STATE{$\bm{x}_E' \leftarrow \bm{x}_E' \cup \{ \tilde{x}^*\}$}
        \ELSE
        \STATE {break}
        \ENDIF
        \ENDWHILE
        \STATE {\bf output:} $\bm{x}_{M''}^* \leftarrow$ $\bm{x}_E'^{\downarrow\bm{X}_M\setminus\bm{X}_M'}$
    \end{algorithmic}
\end{algorithm}

\section{Numerical Experiments}
For a first empirical validation of Algorithms \ref{alg:mar2mmap} and \ref{alg:mar2mmap2}, we consider a benchmark of seven publicly available Markov random fields with different characteristics. Details about these GMs are in Table \ref{tab:bench}, where $n$ denotes the number of model variables, $f$ the number of potentials, and $\omega$ the maximum cardinality of the variables. For each network we generate a random MMAP instance as follows: (i) we select $k$ variables from $\bm{X}$; (ii) we generate a random observation of those variables. Given this evidence, we run Algorithm \ref{alg:mar2mmap2} with an entropy threshold level equal to $\epsilon$ and regard the variables explained by the algorithm after termination as $\bm{X}_M$.\footnote{Code available at \url{https://github.com/Tioz90/MMAP2MAR}.} For both MAR and MMAP queries, we use the state-of-the-art exact solver Merlin built on top of And/Or search as developed in \cite{marinescu2014} and later extended with stochastic search in \cite{marinescu2018stochastic}.\footnote{\url{https://github.com/radum2275/merlin}}

\begin{table}[htp]
\begin{tabular}{clccc}
\hline
Id&Filename&$n$&$f$&$\omega$\\
\hline
(a)&{\tt GEOM30a\_3.wcsp.uai}&$30$&$81$&$3$\\
(b)&{\tt GEOM30a\_4.wcsp.uai}&$30$&$81$&$4$\\
(c)&{\tt rbm\_ferro\_22.uai}&$44$&$528$&$2$\\
(d)&{\tt driverlog01ac.wcsp.uai}&$71$&$618$&$4$\\
(e)&{\tt grid10x10.f10.uai}&$100$&$280$&$2$\\
%(f)&{\tt or\_chain\_111.fg.uai}&$200$&$200$&$2$\\
(f)&{\tt 1502.wcsp.uai}&$209$&$411$&$4$\\
\hline
\end{tabular}
\caption{Markov random fields benchmark\label{tab:bench}}
\end{table}

We denote as $T_{\mathrm{MAR}}$ the cumulative execution time used by the approximate algorithm when calling the MAR tasks, and $T_{\mathrm{MMAP}}$ the execution time required for the exact solution of the MMAP task. The two solutions are compared both in terms of \emph{exact match}, i.e., how many times the exact and the approximate sequence of variables to be explained are equal, and \emph{normalized Hamming similarity}, i.e., one minus the normalized Hamming distance between the two sequences. For each model the procedure is iterated $q$ times and the average values are reported.

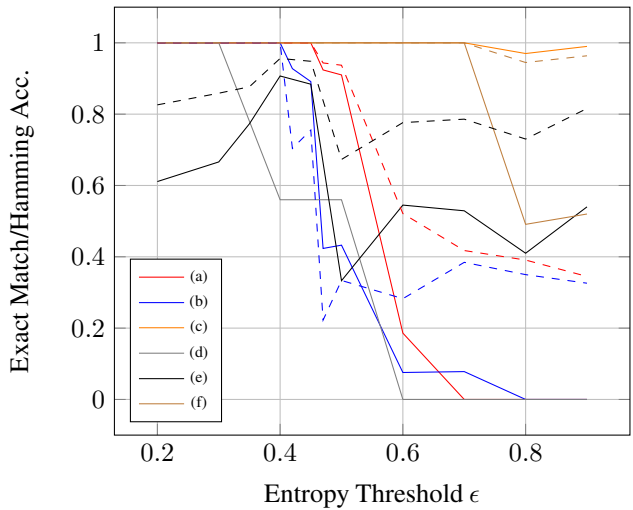
\begin{figure}[htp!]
\centering
    \begin{tikzpicture}[yscale=1,xscale=1]
    \begin{axis}[
    xlabel = {Entropy Threshold $\epsilon$},
    ylabel = {Exact Match/Hamming Acc.},
    %width=1.0\textwidth,
    %height=0.6\textwidth,
    grid = major,
    legend pos = south west,
    legend style={font=\tiny},
    legend entries = {(a),(b),(c),(d),(e),(f)}]
    \addplot[color=red] table {a_5_match.dat};
    \addplot[color=blue] table {b_5_match.dat};
    \addplot[color=orange] table {c_5_match.dat};
    \addplot[color=gray] table {d_5_match.dat};
    \addplot[color=black] table {e_5_match.dat};
    \addplot[color=brown] table {g_5_match.dat};
    \addplot[color=red,dashed] table {a_5_hamming.dat};
    \addplot[color=blue,dashed] table {b_5_hamming.dat};
    \addplot[color=black,dashed] table {e_5_hamming.dat};
    \addplot[color=brown,dashed] table {g_5_hamming.dat};
    \end{axis}
    \end{tikzpicture}
    \caption{Accuracy trajectories for exact match (line) and Hamming (dashed) accuracies with $k=5$ and $q=1000$\label{fig:accuracy}}
\end{figure}

In Figure \ref{fig:accuracy}, we report the exact-match accuracy trajectories (continuous lines) on the benchmark models for increasing values of the entropy threshold. If the exact-match accuracy is not one, also the Hamming accuracy (dashed line) accuracy is reported. As expected both accuracies decrease for increasing values of $\epsilon$. Notably for low entropy thresholds the algorithm reach very high accuracy levels, while the smoother behaviour of the Hamming trajectories shows that accepting variables with higher entropies produces wrong explanations still including many variables in their right state. This basically proves that the quality of a MMAP solution as achieved by Algorithm \ref{alg:mar2mmap} depends on the maximum entropy of the variables explained during the different iterations, and this value can be safely regarded as a confidence level of the quality of the resulting solution.

Regarding the execution times, the slowest exact MMAP inferences has been computed for network (d). Remarkably on those instances MMAP2MAR is two order of magnitude faster and we have the average value $T_{\mathrm{MMAP}}/
T_{\mathrm{MAR}}\simeq 83$. On the other models, exact MMAP inference is fast and the two approaches have the same order of magnitude. Similar results have been obtained for different values of $k$.

\section{Conclusions and Outlooks}
We presented a heuristic approach to MMAP inference in probabilistic graphical models (both Bayesian networks and Markov random field). The algorithm reduces such a NP$^\mathrm{PP}$-complete task to a polynomial number of marginalizations of single variables. Despite its simplicity, preliminary experiments show surprisingly accurate result in finding the most probable explanation when the reliability measure defined together with the algorithm is high. As a future work we intend to provide a deeper experimental validation and also evaluate the possibility of an application of this scheme as a general XAI tool for graphical models, this being in line with the ideas  originally presented in \cite{tiotto2019}.

\bibliographystyle{aaai}
\bibliography{biblio}
\end{document}